\pdfoutput=1

\documentclass[11pt]{article}

\usepackage[final]{EACL2023}

\usepackage{times}
\usepackage{latexsym}

\usepackage[T1]{fontenc}

\usepackage[utf8]{inputenc}

\usepackage{microtype}

\usepackage{inconsolata}

\usepackage{amsmath}
\usepackage{mathrsfs}
\usepackage{graphics}
\usepackage{url}
\usepackage{xr}
\usepackage{multirow}
\usepackage{tabularx}
\usepackage{booktabs}
\usepackage{balance}
\usepackage{caption}
\usepackage{subcaption}
\usepackage{enumitem}
\usepackage{pgfplots}
\pgfplotsset{compat=1.17}
\usepackage{tikz}
\usepackage{xcolor}
\usepackage{xspace}
\usepackage{makecell}
\usepackage{amssymb}
\usepackage{graphicx}
\usepackage{calc}

\newcommand{\bert}[1]{\textsc{BERT}$_{#1}$\xspace}
\newcommand{\electra}[1]{\textsc{ELECTRA}$_{#1}$\xspace}
\newcommand{\nfr}{$\mathcal{R}_{NF}$\xspace}

\title{Improving Prediction Backward-Compatiblility in NLP Model Upgrade with Gated Fusion}

\author{Yi-An Lai$^{\spadesuit}$\quad Elman Mansimov$^{\spadesuit}$\quad Yuqing Xie$^{\heartsuit,}$\thanks{~~Work done during author's internship at AWS AI Labs.}~\quad Yi Zhang$^{\spadesuit}$ \\
  $^{\spadesuit}$AWS AI Labs
  \quad $^{\heartsuit}$University of Waterloo \\
  \texttt{\{yianl,mansimov,yizhngn\}@amazon.com} \\
  \texttt{yuqing.xie@uwaterloo.ca} \\
}

\begin{document}
\maketitle
\begin{abstract}
When upgrading neural models to a newer version, new errors that were not encountered in the legacy version can be introduced, known as \emph{regression\footnotemark errors}.
This inconsistent behavior during model upgrade often outweighs the benefits of accuracy gain and hinders the adoption of new models.
To mitigate regression errors from model upgrade, distillation and ensemble have proven to be viable solutions without significant compromise in performance.
Despite the progress, these approaches attained an incremental reduction in regression which is still far from achieving backward-compatible model upgrade. 
In this work, we propose a novel method, \emph{Gated Fusion}, that promotes backward compatibility via learning to mix predictions between old and new models.
Empirical results on two distinct model upgrade scenarios show that our method reduces the number of regression errors by 62\% on average, outperforming the strongest baseline by an average of 25\%.
\footnotetext{Within this work, \textit{regression} denotes performance degradation in software systems, instead of the statistical technique for estimating relationships among variables.}
\end{abstract}

\section{Introduction}

\begin{figure}[t!]
\centering
\includegraphics[width=0.4\textwidth]{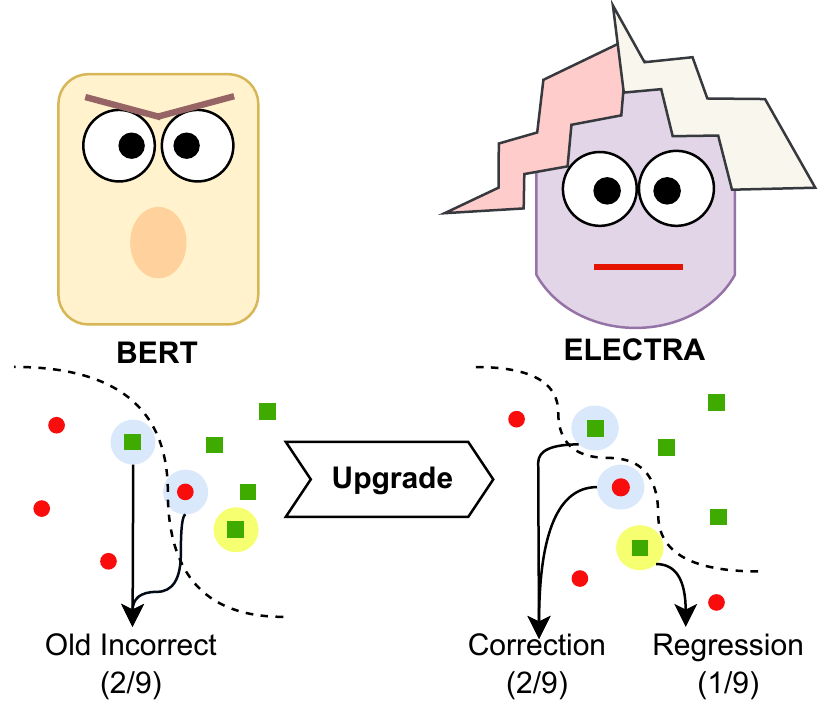}
\caption{Illustration of regression errors when upgrading from BERT \citep{devlin-etal-2019-bert} to ELECTRA \citep{Clark2020ELECTRA:} for classification.
Red circles and green squares denote examples of different classes. 
Dashed lines represent decision boundaries.
}
\label{fig:regression}
\end{figure}

In order to achieve a smooth continuous improvement of NLP applications, it is critical to guarantee consistent operation of the system after an upgrade.
New errors introduced during the model upgrade interfere with the existing user experience and are considered to be a \emph{regression} in the quality.
Due to the difficulty of modularizing or explaining the behavior of deep neural networks, traditional software regression tests are inapplicable to neural based systems. 
The cost of arduous error analysis and model patching often exceeds the benefits of model upgrades.
Developing methods that ensure backward compatibility during model upgrades without compromise in performance becomes a valuable research direction \citep{yan2021positive,work-in-progress,trauble2021backward,cai2022measuring}.

The \textit{prediction backward-compatible model upgrade} problem aims to improve consistency of correct classification predictions between legacy and upgraded models without accuracy loss.
\citet{yan2021positive} first studied backward compatibility during model upgrade on image classification tasks. They proposed to enforce the positive congruence of the new model with the old one by applying a knowledge distillation objective \citep{Hinton2015DistillingTK} objective with re-weighting of training samples.
Later, \citet{work-in-progress} extended the work of \citet{yan2021positive} by investigating the backward compatibility in NLP classification tasks. They found that their proposed distillation-based approach can help decrease the regression errors of specific linguistic phenomena in NLP classification tasks.

Despite progress with both distillation- and ensemble-based regression-mitigation approaches, there are limitations that prevent them from broad practical adoption in ML operations.
Distillation-based methods attempt to transfer the prediction power of the old model to the new one on potential regression instances \citep{Hinton2015DistillingTK}.
However, given the huge complexity of current neural architectures and relatively scarce training data in downstream tasks, models could have insufficient data to reliably estimate the probable regression cases and carry out the transfer on them \citep{work-in-progress,cai2022measuring}.
On the other hand, model ensemble aggregates predictions from differently-trained new models but bears no connection with the legacy version \citep{yan2021positive}.
These limitations reveal the two major challenges when striving to ensure backward compatibility.
First, the new model could have distinct inductive bias and prediction behavior than the old system, rooted from inherent differences such as architecture, model size, and pretraining procedure \citep{liu2021self}.
Second, during new model training, a reliable mechanism is needed in place to bridge the gap between two models and mitigate potential inconsistencies. 

Inspired by the strength and weakness of prior approaches, we propose \emph{Gated Fusion} to integrate old and new models via gating mechanism \citep{hochreiter1997long,chung2014empirical,gu2016incorporating}, essentially a light-weight ensemble of legacy and upgrade models connected via a learned fusion gate.
Specifically, we add a \textit{learned} gate on top of the new model. We combine the logits from old and new models according to the weight from the gate. We train our Gated Fusion model by minimizing the standard cross-entropy error.
The intuition is that the gate could learn to put more weights on the old model when the new model cannot produce correct predictions, effectively doing fall-backs that optimizes backward compatibility.

Empirical results demonstrate that our proposed approach outperforms other competing methods significantly, where we can obtain on average $62\%$ reduction of total negative flips, i.e. new errors caused by the model upgrade, without any degradation in accuracy performance.
The effectiveness of Gated Fusion is validated across three diverse classification tasks and two distinct model upgrade scenarios (a) upgrade to larger model size (b) upgrade to advanced pretrained model, where consistent results are attained across the board.

Our main contributions are as follows:
\begin{itemize}[itemsep=4pt,topsep=4pt,parsep=0pt,partopsep=0pt]
    \item We propose Gated Fusion that integrates old and new models via gating mechanism for backward-compatible model upgrade;
    \item We evaluate competing methods on two distinct and challenging model upgrade scenarios across three diverse classification tasks;
    \item Empirical results show that our proposed approach significantly outperforms competing methods and achieves regression reductions by a large margin across the board.
\end{itemize}

\section{The Backward-Compatible Model Upgrade Problem}

\begin{figure*}[t!]
\centering
\includegraphics[width=0.9\textwidth]{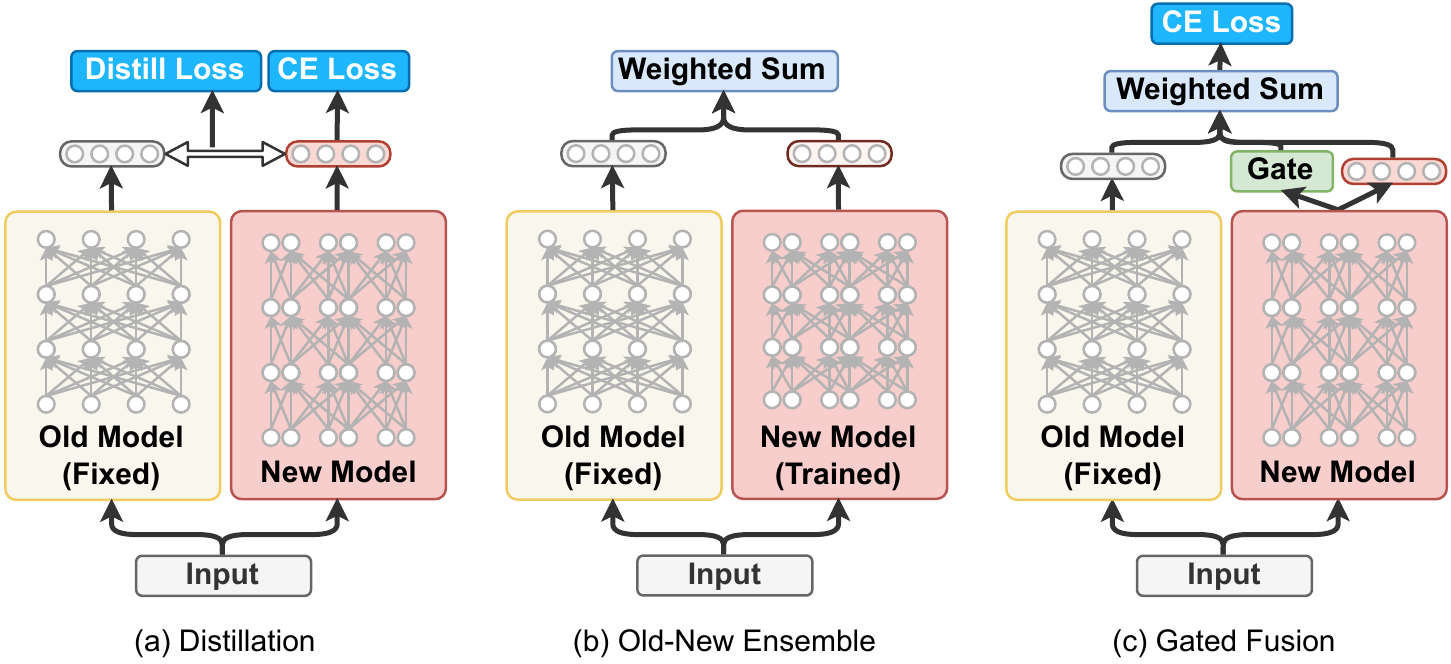}
\caption{Methods to improve prediction backward compatibility during model upgrade. (a) Distillation-based approach to align predicted logits on potential regression instances \citep{work-in-progress}. (b) Ensemble of old and new models via weighted sum of either predicted logits or probabilities. (c) Our proposed Gated Fusion that learns a gate as a soft switch to dynamically determine whether to fall back to previous predictions.}
\label{fig:method}
\end{figure*}

The goal of backward-compatible model upgrade is to minimize regression errors without compromising the accuracy performance during model upgrade \citep{yan2021positive,work-in-progress}.
In this work, we aim to improve the backward compatibility of model predictions in the NLP classification tasks. Following \citet{work-in-progress}, we study the scenario where the underlying pretrained language model (LM) is being upgraded.

Let $x$ be a natural language input with a class label $y \in \{1, 2, ..., C\}$. $\mathcal{D}=\{x_i, y_i\}_{i=1}^{N}$ denotes a set of $N$ examples with corresponding labels.
A classifier $f$ estimates the class probabilities given the input
$\vec{f}(x) = (p(y=1|x), ..., p(y=C|x))^{\top}$.
When upgrading from an \emph{old} model $f_{old}$ to a \emph{new} model $f_{new}$, normally with distinct architectures and trained on the same data, an improved model $f^{*}$ is produced based on $f_{old}$ and $f_{new}$.
Our goal is for $f^{*}$ to minimize \emph{regression errors} as an additional objective, while still achieving comparable performance to $f^{o}_{new}$, the new model trained in the vanilla setting.
Note that $f^{*}$ could be multiple times larger than $f^{o}_{new}$, with model ensemble of $f^{o}_{new}$ as one example \citep{yan2021positive}.

\vspace{-0.05cm}
\paragraph{Measuring Backward Compatibility.}
The backward compatibility is measured via quantifying regression errors on a given regression measurement set $\mathcal{D}_{reg}=\{x_i, y_i\}_{i=1}^{M}$.
$\mathcal{D}_{reg}$ could be a hidden customer test set comprising critical use cases, a set of behavioral testing examples for targeted evaluation \citep{ribeiro-etal-2020-beyond}, or the development split from the dataset of interest.
In this work, we take the development set as our $\mathcal{D}_{reg}$ for evaluation.

For classification, regression errors are characterized by \emph{negative flips}, denoted as $\mathcal{R}_{NF}$ -- the portion of samples in $\mathcal{D}_{reg}$ that flip from correct prediction $f_{old}(x_i)=y_i$ to incorrect output $f_{new}(x_i)\neq y_i$ during model upgrade:
\begingroup\makeatletter\def\f@size{10}\check@mathfonts
\begin{equation}
\mathcal{R}_{NF}(\mathcal{D}_{reg}, \vec{f}_{old}, \vec{f}_{new})
=\frac{|\{x|f_{old}=y, f_{new}\neq y\}|}
{|\mathcal{D}_{reg}|}.
\end{equation}
\endgroup
One thing to emphasize is that maximizing classifier performance does not necessarily help in minimizing \nfr\citep{yan2021positive,work-in-progress}.

\section{Gated Fusion: Methodology}

\subsection{Method Overview} \label{model}

To improve backward compatibility in model upgrade, it's crucial to have a mechanism that detects potential regression errors and mitigates them when making predictions.
We propose Gated Fusion (GF) to achieve this by learning a gate as a soft switch to choose between generating predictions by the new model or resorting to outputs of the old model. 
Gated Fusion is inspired by the gating mechanism widely used in other applications. For example, mixing word copying mode with word generation mode for language modeling \citep{merity2016pointer} and summarization \citep{see2017get}.

Our proposed Gated Fusion $f^{*}_{GF}$ consists of the old model $f_{old}$, the new model $f_{new}$, and a gating network $g_{\theta}$.
The old model $f_{old}$ is the legacy version before upgrade where the parameters are fixed.
The new model $f_{new}$ has the same architecture as $f^{o}_{new}$ and is randomly initialized. 
The gating network $g_{\theta}$ is a multi-layer feed-forward network with sigmoid function. It produces a scalar weight $\alpha_{gate}$ in the range $[0, 1]$ from the output layer of $f_{new}$, denoted as $\mathcal{E}_{new}$:
\begingroup\makeatletter\def\f@size{10}\check@mathfonts
\begin{equation}
\alpha_{gate}(x) = g_{\theta}(\mathcal{E}_{new}(x)).
\end{equation}
\endgroup
\noindent We use $\alpha_{gate}$ to combine the logits of old and new models as our final outputs:
\begingroup\makeatletter\def\f@size{10}\check@mathfonts
\begin{equation}
l^{*}_{GF}(y|x)
= ( 1-\alpha_{gate}) \cdot \frac{l_{old}(y|x)}{T} + \alpha_{gate} \cdot l_{new}(y|x),
\end{equation}
\endgroup
where $l(y|x)$ denotes predicted logits from models and $T$ is the temperature scaling to regularize the magnitude of old model's logits.
$f_{new}$ and $g_{\theta}$ are then jointly trained end-to-end with cross-entropy loss between our output logits $l^{*}_{GF}(y|x)$ and label distributions on downstream tasks.

The intuition behind Gated Fusion is that when $f_{new}$ makes a mistake while $f_{old}$ produces the correct output, the gate $g_{\theta}$ will learn to put more weight on $f_{old}$ in order to minimize the final classification loss.
This process effectively mitigates potential negative flips introduced by the model upgrade and thus improves the backward compatibility of final predictions.


\subsection{Training and Inference}

In practice, training Gated Fusion with randomly initialized $f_{new}$ would make the shallow gating network quickly converge to favor the fully-trained $f_{old}$.
To prevent this, we only train $f_{new}$ for the first few epochs to ensure its competence before jointly training $g_{\theta}$ and $f_{new}$ using $l^{*}_{GF}(x)$.
In addition, we found that stopping gradient flow from $g_{\theta}$ to $f_{new}$ can further prevent the performance decrease of the new model within Gated Fusion:
\begingroup\makeatletter\def\f@size{10}\check@mathfonts
\begin{equation}
\alpha_{gate}(x) = g_{\theta}(\mathit{stop\_grad}(\mathcal{E}_{new}(x))).
\end{equation}
\endgroup

At inference time, Gated Fusion produces logits from $f_{old}$ and $f_{new}$ as well as the gate value $\alpha_{gate}$ to make output predictions:
\begingroup\makeatletter\def\f@size{10}\check@mathfonts
\begin{equation}
f^{*}_{GF}(x)
= \mathit{Softmax}\Big( (1-\alpha_{gate}) \cdot \frac{l_{old}}{T} + \alpha_{gate} \cdot l_{new}\Big).
\end{equation}
\endgroup

\subsection{Inference with Cache}\label{cache}

Our proposed Gated Fusion requires $f_{old}$ to be hosted together with the new model.
In reality, one could have a resource-constrained setting and request the old model to be discarded at inference.
We note that in real applications, repetitive inputs are commonly seen in live traffic \citep{batrinca2015social} and the backward compatibility of model upgrade entails that correct predictions can be preserved on the legacy instances already seen and predicted by the old model.

To simulate real scenarios, we randomly cache old model's logits on a portion of test inputs.
When getting out-of-cache instances, we use new model's output embedding $\mathcal{E}_{new}(x)$ as key and euclidean distance as metric to search for the nearest cached instance.
The cached old-model logits can then be used for Gated Fusion to make predictions without hosting $f_{old}$ at inference.

\section{Experiments Setup}

\subsection{Model Upgrade Scenarios}

We conduct experiments on two representative model upgrade scenarios: 
(a) upgrade to a larger pretrained model of the same type, where we use \emph{\bert{base} to \bert{large}}.   
(b) upgrade to a distinct pretrained model with the same size.
We use \emph{\bert{base} to \electra{base}} \citep{Clark2020ELECTRA:} as this challenging model upgrade scenario for backward-compatibility, as they are pretrained under different self-supervised learning paradigms.
The former uses masked language modeling (MLM) with reconstruction loss, while the latter is pretrained in generative-contrastive (adversarial) fashion with distributional divergence as the loss \citep{liu2021self}.

\subsection{Datasets and Implementation}

We evaluate our approach across three datasets.
They represent different sentence-level classification tasks, from single-sentence to sentence-pair classification, with varying dataset sizes. We use:
(a) Stanford Sentiment Treebank (SST-2), a single-sentence task to classify movie review sentiment, with $67$k train and $0.9$k dev set \citep{socher2013recursive}.
(b) Microsoft Research Paraphrase Corpus (MRPC) \citep{dolan2005automatically}, a sentence-pair classification task for identifying paraphrases, with $3.7$k train and $0.4$k dev set.
(c) Question Natural Language Inference (QNLI), a question-paragraph pair task to determine whether the paragraph contains the answer to the question, with $100$k train and $5.5$k dev set.
Datasets are taken from GLUE Benchmark \citep{wang2018glue} and processed with scripts from Hugging Face\footnote{\scriptsize \url{https://huggingface.co/datasets/glue}}.

For implementation, we use the sequence classification and pre-trained model parameters from Hugging Face Transformers\footnote{\scriptsize \url{https://huggingface.co/docs/transformers/index}}.
Experiments are done in PyTorch \citep{NEURIPS2019_9015} with Tesla V100 GPUs and results are averaged over $5$ random seeds.
Learning rate, batch size, and train epoch are tuned during training new model alone on given tasks and then fixed for all backward-compatible solutions.
In Gated Fusion, we first train $f_{new}$ alone for first $(N - 1)$ epochs and then jointly train $g_{\theta}$ and $f_{new}$ with Gated Fusion logits $l^{*}_{GF}$ in the last epoch.
Further implementation details can be found in the Appendix.

\subsection{Baselines}

We compare our approach with several strong baselines.
(a) Train the new model directly on the target task without any adjustment, i.e. $f^{o}_{new}$. 
(b) The specialized distillation method proposed in \citet{work-in-progress}, where the KL-divergence of prediction probabilities between old and new models is applied when $p_{old}(y=y_i|x_i) > p_{new}(y=y_i|x_i)$.
(c) Model ensemble via majority-voting that was shown to be very effective \citep{yan2021positive,work-in-progress}.
Similarly, we use $5$-seed new model ensemble as a strong baseline.
(d) The ensemble of the old and new models probabilities, $p^{*}(y|x)=(1-\alpha) \cdot p_{old}(y|x) + \alpha \cdot p_{new}(y|x)$, as well as ensemble of the old and new models logits, $l^{*}(y|x)=(1-\alpha) \cdot l_{old}(y|x) + \alpha \cdot l_{new}(y|x)$, where $\alpha$ is searched among $[0.5, 0.6, 0.7, 0.8, 0.9]$ to maximize backward-compatibility while achieving accuracy on par with the vanilla $f^{o}_{new}$.

\section{Results and Analysis}

\begin{table*}[t!]
\centering
\resizebox{0.9\textwidth}{!}{
\begin{tabular}{l|cc|cc|cc}
\toprule
& 
\multicolumn{2}{c|}{\textbf{SST-2}} &
\multicolumn{2}{c|}{\textbf{MRPC}} & 
\multicolumn{2}{c}{\textbf{QNLI}}\\
\textbf{\bert{base} $\rightarrow$ \bert{large}} & \nfr & Accuracy & \nfr & Accuracy & \nfr & Accuracy \\
\midrule
Old Model & 
- & 92.00$_{\tiny 0.27}$ & 
- & 85.69$_{\tiny 0.90}$ &
- & 90.74$_{\tiny 0.09}$ \\
New Model &
2.18$_{\tiny 0.21}$ & 93.12$_{\tiny 0.29}$ &
4.12$_{\tiny 1.04}$ & 87.40$_{\tiny 1.02}$ &
2.72$_{\tiny 0.13}$ & 92.22$_{\tiny 0.16}$ \\
\midrule
Distillation \citep{work-in-progress} &
1.97$_{\tiny 0.22}$ & 93.33$_{\tiny 0.20}$ &
3.53$_{\tiny 0.77}$ & 87.70$_{\tiny 1.34}$ &
2.31$_{\tiny 0.14}$ & 92.60$_{\tiny 0.19}$ \\
New Model Ensemble &
2.00$_{\tiny 0.31}$ & 93.30$_{\tiny 0.24}$ &
2.25$_{\tiny 0.61}$ & 88.87$_{\tiny 0.77}$ &
1.98$_{\tiny 0.21}$ & 92.97$_{\tiny 0.22}$ \\
Old-New Probs Ensemble & 
1.06$_{\tiny 0.27}$ & 93.12$_{\tiny 0.38}$ &
1.67$_{\tiny 0.78}$ & 87.16$_{\tiny 1.12}$ &
1.04$_{\tiny 0.26}$ & 92.44$_{\tiny 0.23}$ \\
Old-New Logits Ensemble &
1.06$_{\tiny 0.27}$ & 93.12$_{\tiny 0.38}$ &
1.67$_{\tiny 0.78}$ & 87.16$_{\tiny 1.12}$ &
1.04$_{\tiny 0.26}$ & 92.44$_{\tiny 0.23}$ \\
Gated Fusion &
\textbf{0.78}$_{\tiny 0.20}$ & 93.05$_{\tiny 0.09}$ & 
\textbf{1.18}$_{\tiny 0.52}$ & 87.45$_{\tiny 0.52}$ &
\textbf{0.73}$_{\tiny 0.13}$ & 92.24$_{\tiny 0.24}$ \\
\bottomrule
\end{tabular}
}
\caption{Negative flip rate \nfr and model accuracy ($\%$) of competing methods to optimize backward-compatibility without performance degradation during \bert{base} $\rightarrow$ \bert{large} model upgrade.}
\label{tab:bertlarge_main}
\end{table*}

\begin{table*}[t!]
\centering
\resizebox{0.9\textwidth}{!}{
\begin{tabular}{l|cc|cc|cc}
\toprule
& 
\multicolumn{2}{c|}{\textbf{SST-2}} &
\multicolumn{2}{c|}{\textbf{MRPC}} & 
\multicolumn{2}{c}{\textbf{QNLI}}\\
\textbf{\bert{base} $\rightarrow$ \electra{base}} & \nfr & Accuracy & \nfr & Accuracy & \nfr & Accuracy \\
\midrule
Old Model & 
- & 92.00$_{\tiny 0.27}$ & 
- & 85.69$_{\tiny 0.90}$ &
- & 90.74$_{\tiny 0.09}$ \\
New Model &
1.63$_{\tiny 0.20}$ & 95.00$_{\tiny 0.06}$ &
3.73$_{\tiny 0.36}$ & 88.58$_{\tiny 0.57}$ &
2.82$_{\tiny 0.32}$ & 92.90$_{\tiny 0.26}$ \\
\midrule
Distillation \citep{work-in-progress} &
1.49$_{\tiny 0.24}$ & 95.02$_{\tiny 0.21}$ &
3.68$_{\tiny 0.79}$ & 88.82$_{\tiny 0.94}$ &
2.58$_{\tiny 0.17}$ & 93.03$_{\tiny 0.16}$ \\
New Model Ensemble &
1.12$_{\tiny 0.09}$ & 95.39$_{\tiny 0.09}$ &
3.24$_{\tiny 0.24}$ & 89.02$_{\tiny 0.48}$ &
2.26$_{\tiny 0.08}$ & 93.49$_{\tiny 0.07}$ \\
Old-New Probs Ensemble & 
1.40$_{\tiny 0.17}$ & 95.07$_{\tiny 0.15}$ &
3.14$_{\tiny 0.42}$ & 88.53$_{\tiny 0.48}$ &
0.98$_{\tiny 0.20}$ & 93.04$_{\tiny 0.21}$ \\
Old-New Logits Ensemble &
0.89$_{\tiny 0.17}$ & 94.95$_{\tiny 0.13}$ &
3.28$_{\tiny 0.43}$ & 88.48$_{\tiny 0.51}$ &
0.98$_{\tiny 0.20}$ & 93.04$_{\tiny 0.21}$ \\
Gated Fusion &
\textbf{0.71}$_{\tiny 0.18}$ & 95.02$_{\tiny 0.16}$ & 
\textbf{2.40}$_{\tiny 0.50}$ & 88.68$_{\tiny 0.68}$ &
\textbf{0.81}$_{\tiny 0.16}$ & 92.98$_{\tiny 0.17}$ \\
\bottomrule
\end{tabular}
}
\caption{Negative flip rate \nfr and model accuracy ($\%$) of competing methods to optimize backward-compatibility without performance degradation during \bert{base} $\rightarrow$ \electra{base} model upgrade.}
\label{tab:electra_main}
\end{table*}

\subsection{Upgrade to a Larger Pretrained Model}

Our first model upgrade scenario scales up the size of underlying pretrained language models. 
We experiment with \bert{base} to \bert{large}, where the model size is tripled (110M vs 340M) and the model depth is doubled (12 vs 24 layers).

Table \ref{tab:bertlarge_main} shows the results.
For $f^{o}_{new}$, we can observe that the negative flip rates \nfr are usually much larger than the accuracy gains across tasks, which could be the reason to hinder new model adoptions in real-world applications.
Besides, when dividing \nfr over the error rate $(1 - \textit{accuracy})$, we can observe that around $30\%$ to $40\%$ of all $f^{o}_{new}$ prediction errors are in fact the \emph{new} errors introduced during model upgrade.
For improving prediction backward-compatibility, our proposed Gated Fusion outperforms other competing methods to considerably reduce \nfr without degradation on accuracy.
Note that best $\alpha$ values found for the two variants of old-new ensemble are both $0.5$, hence producing identical results.

Compared to the vanilla new model, gated fusion obtains absolute \nfr reductions of \textminus$1.40\%$ on SST-2, \textminus$2.94\%$ on MRPC, and \textminus$1.99\%$ on QNLI.
These translate to reducing the total negative flip cases by $64.2\%$, $71.4\%$, $73.2\%$, respectively.
Compared to the strongest baseline (old-new ensemble), we obtain further absolute \nfr reductions of \textminus$0.28\%$ on SST-2, \textminus$0.49\%$ on MRPC, and \textminus$0.31\%$ on QNLI, which translate to further reducing $12.8\%$, $11.9\%$, and $11.4\%$ of negative flip cases.
These results show the effectiveness of our method to mitigate a significant amount of regression errors during model upgrade.

\begin{table*}[t!]
\centering
\resizebox{0.85\textwidth}{!}{
\begin{tabular}{l|cc|cc|cc}
\toprule
&
\multicolumn{2}{c|}{\textbf{SST-2}} &
\multicolumn{2}{c|}{\textbf{MRPC}} & 
\multicolumn{2}{c}{\textbf{QNLI}}\\
 & \nfr & Accuracy & \nfr & Accuracy & \nfr & Accuracy \\
\midrule
Old Model: \bert{base} & 
- & 92.00$_{\tiny 0.27}$ & 
- & 85.69$_{\tiny 0.90}$ &
- & 90.74$_{\tiny 0.09}$ \\
\midrule
New Model: \bert{large} &
2.18$_{\tiny 0.21}$ & 93.12$_{\tiny 0.29}$ &
4.12$_{\tiny 1.04}$ & 87.40$_{\tiny 1.02}$ &
2.72$_{\tiny 0.13}$ & 92.22$_{\tiny 0.16}$ \\
Model Ensemble: 5 seeds &
2.00$_{\tiny 0.31}$ & 93.30$_{\tiny 0.24}$ &
2.25$_{\tiny 0.61}$ & 88.87$_{\tiny 0.77}$ &
1.98$_{\tiny 0.21}$ & 92.97$_{\tiny 0.22}$ \\
Model Ensemble: 10 seeds &
1.79$_{\tiny 0.17}$ & 93.69$_{\tiny 0.15}$ &
2.01$_{\tiny 0.29}$ & 89.46$_{\tiny 0.51}$ &
2.01$_{\tiny 0.14}$ & 92.97$_{\tiny 0.19}$ \\
Model Ensemble: 20 seeds &
1.79$_{\tiny 0.25}$ & 93.62$_{\tiny 0.16}$ &
1.76$_{\tiny 0.50}$ & 89.56$_{\tiny 0.48}$ &
1.82$_{\tiny 0.15}$ & 93.13$_{\tiny 0.08}$ \\
Gated Fusion &
0.78$_{\tiny 0.20}$ & 93.05$_{\tiny 0.09}$ & 
1.18$_{\tiny 0.52}$ & 87.45$_{\tiny 0.52}$ &
0.73$_{\tiny 0.13}$ & 92.24$_{\tiny 0.24}$ \\
\midrule
New Model: \electra{base} &
1.63$_{\tiny 0.20}$ & 95.00$_{\tiny 0.06}$ &
3.73$_{\tiny 0.36}$ & 88.58$_{\tiny 0.57}$ &
2.82$_{\tiny 0.32}$ & 92.90$_{\tiny 0.26}$ \\
Model Ensemble: 5 seeds &
1.12$_{\tiny 0.09}$ & 95.39$_{\tiny 0.09}$ &
3.24$_{\tiny 0.24}$ & 89.02$_{\tiny 0.48}$ &
2.26$_{\tiny 0.08}$ & 93.49$_{\tiny 0.07}$ \\
Model Ensemble: 10 seeds &
1.24$_{\tiny 0.18}$ & 95.30$_{\tiny 0.16}$ &
3.63$_{\tiny 0.50}$ & 88.58$_{\tiny 0.20}$ &
2.21$_{\tiny 0.12}$ & 93.57$_{\tiny 0.15}$ \\
Model Ensemble: 20 seeds & 
1.19$_{\tiny 0.16}$ & 95.32$_{\tiny 0.17}$ &
3.43$_{\tiny 0.51}$ & 88.92$_{\tiny 0.48}$ &
2.15$_{\tiny 0.17}$ & 93.63$_{\tiny 0.11}$ \\
Gated Fusion &
0.71$_{\tiny 0.18}$ & 95.02$_{\tiny 0.16}$ &
2.40$_{\tiny 0.50}$ & 88.68$_{\tiny 0.68}$ &
0.81$_{\tiny 0.16}$ & 92.98$_{\tiny 0.17}$ \\
\bottomrule
\end{tabular}
}
\caption{Negative flip rate \nfr and model accuracy ($\%$) when increasing number of seeds used in new model ensemble, comparing with our proposed method (Gated Fusion).}
\label{tab:ensemble}
\end{table*}

\vspace{-0.03cm}
\subsection{Upgrade to a Different Pretrained Model}
\vspace{-0.03cm}
A more challenging upgrade scenario is when old and new models are pretrained under distinctive paradigms, producing two representation spaces of fairly different characteristics \citep{meng2021coco}.
We experiment with \bert{base} to \electra{base} in this scenario, where two models have the same size but are pretrained under utterly different schemes, i.e. generative versus adversarial.

Table \ref{tab:electra_main} shows the results.
For $f^{o}_{new}$, compared with upgrading to \bert{large}, we observe larger accuracy gains and lower \nfr on SST-2 and MRPC.
However, on QNLI, upgrading to \electra{base} achieves a higher accuracy gain but an even a higher \nfr.
This implies that boosting accuracy and improving backward compatibility could be two related but different objectives.

For mitigation strategies, Gated Fusion achieves the lowest negative flip rates across datasets without any accuracy loss.
We obtain absolute \nfr reductions of \textminus$0.92\%$ on SST-2, \textminus$1.33\%$ on MRPC, and \textminus$2.01\%$ on QNLI over the vanilla setup, reducing $56.4\%$, $35.7\%$, and $71.3\%$ of overall negative flips, respectively.
Compared with upgrading to \bert{large}, we observe that upgrading to \electra{base} has much smaller relative negative flip reductions on SST-2 and MRPC, showing that it could be indeed harder to improve backward-compatibility when upgrading to a distinct pretrained model.
In contrast, similar relative negative flip reductions are observed on QNLI across two upgrade scenarios.
This could be attributed to the abundant training data of the downstream task.


\begin{table}[t!]
\centering
\resizebox{0.48\textwidth}{!}{
\begin{tabular}{l|ccc}
\toprule
& 
\textbf{SST-2} & \textbf{MRPC} & \textbf{QNLI}\\
\midrule
Old: \bert{base} & 
92.00$_{\tiny 0.27}$ & 
85.69$_{\tiny 0.90}$ &
90.74$_{\tiny 0.09}$ \\
\midrule
to \bert{large} &
93.12$_{\tiny 0.29}$ &
87.40$_{\tiny 1.02}$ &
92.22$_{\tiny 0.16}$ \\
Gated Fusion &
93.05$_{\tiny 0.09}$ & 
87.45$_{\tiny 0.52}$ &
92.24$_{\tiny 0.24}$ \\
\hspace{0.35cm}$\text{-}$ drop old model &
93.17$_{\tiny 0.61}$ &
87.75$_{\tiny 1.14}$ &
92.22$_{\tiny 0.44}$ \\
\midrule
to \electra{base}&
95.00$_{\tiny 0.06}$ &
88.58$_{\tiny 0.57}$ &
92.90$_{\tiny 0.26}$ \\
Gated Fusion &
95.02$_{\tiny 0.16}$ & 
88.68$_{\tiny 0.68}$ &
92.98$_{\tiny 0.17}$ \\
\hspace{0.35cm}$\text{-}$ drop old model &
95.16$_{\tiny 0.09}$ & 
88.63$_{\tiny 0.94}$ &
93.06$_{\tiny 0.13}$ \\
\bottomrule
\end{tabular}
}
\caption{Accuracy ($\%$) when dropping the old model within Gated Fusion at inference time.}
\label{tab:newalone}
\end{table}

\begin{table}[t!]
\centering
\resizebox{0.47\textwidth}{!}{
\begin{tabular}{l|cc}
\toprule
& 
\multicolumn{2}{c}{\textbf{SST-2}} \\
 & \nfr & Accuracy \\
\midrule
Old Model: \bert{base} & 
- & 92.00$_{\tiny 0.27}$ \\
\midrule
New Model: \electra{base}&
1.63$_{\tiny 0.20}$ & 95.00$_{\tiny 0.06}$ \\
Gated Fusion - 50$\%$ cache &
1.26$_{\tiny 0.10}$ & 94.86$_{\tiny 0.27}$ \\
Gated Fusion - 75$\%$ cache &
0.99$_{\tiny 0.25}$ & 94.91$_{\tiny 0.12}$ \\
Gated Fusion &
0.71$_{\tiny 0.18}$ & 95.02$_{\tiny 0.16}$ \\

\bottomrule
\end{tabular}
}
\caption{Negative flip rate \nfr and model accuracy ($\%$) of Gated Fusion with $X\%$ cache of old model logits at inference time.}
\label{tab:cache}
\end{table}

\subsection{Drop Old Model at Inference Time}

Our proposed method requires the old model to be hosted together with the new model.
A natural question is whether we could train Gated Fusion with the old model and then discard it at inference time to host the new model only.

We first experiment with directly dropping the old model within Gated Fusion at inference time.
Results in Table \ref{tab:newalone} show that dropping old model in Gated Fusion can still achieve comparable accuracy across the board, suggesting no performance degradation.
Nonetheless, we observe that the negative flip rates also fall back to similar positions as training the new model in the vanilla setting.

However, in real application scenario, live inputs are often repetitively seen across time and ensuring backward-compatibility means that correct predictions on same instances can be preserved after model upgrade.
We experiment with the caching method introduced in section \ref{cache} to store old model's logits on random $X\%$ of test instances where Gated Fusion can later access them for inference.
Results in Table \ref{tab:cache} show that with higher percentage of cache, \nfr is gradually reduced towards \nfr of the original Gated Fusion, which is equivalent to $100\%$ cache.
Still, we observe a notable gap in \nfr between the partial caching and full settings. 
We leave the examination of ways to achieve the upper bound in reduction in \nfr with smaller cache to the future work.

\begin{table*}[t!]
\centering
\resizebox{0.96\textwidth}{!}{
\begin{tabular}{ccp{15cm}}
\toprule
&\textbf{(Task, Label)} & \textbf{Examples}\\
\midrule\vspace{0.2cm}
\parbox[t]{2mm}{\multirow{9}{*}{\rotatebox[origin=c]{90}{\textbf{\bert{base} $\rightarrow$ \bert{large}}}}}
& (SST-2, Positive) & [Sentence] A study in shades of gray, offering itself up in subtle plot maneuvers ... \\\vspace{0.1cm}
& (SST-2, Negative) & [Sentence] Manages to be both repulsively sadistic and mundane. \\\vspace{0.1cm}
& (MRPC, Not Equivalent) & [Sentence 1] Vivace was founded in 1999 and has raised over \$118 million in three rounds of venture financing. [Sentence 2] During difficult times for technology venture capital, Vivace raised over \$118 million in three rounds of venture financing. \\\vspace{0.1cm}
& (QNLI, Entailment) & [Question] Why was there a depreciation of the industrialized nations dollars? [Sentence] Anticipating that currency values would fluctuate unpredictably for a time, the industrialized nations increased their reserves (by expanding their money supplies) in amounts far greater than before. \\\midrule
\parbox[t]{2mm}{\multirow{12}{*}{\rotatebox[origin=c]{90}{\textbf{\bert{base} $\rightarrow$ \electra{base}}}}}
& (SST-2, Positive) & [Sentence] Aside from minor tinkering , this is the same movie you probably loved in 1994, except that it looks even better. \\\vspace{0.1cm}
& (SST-2, Negative) & [Sentence] It showcases carvey's talent for voices, but not nearly enough and not without taxing every drop of one's patience to get to the good stuff . \\\vspace{0.1cm}
& (MRPC, Equivalent) & [Sentence 1] Blair's Foreign Secretary Jack Straw was to take his place on Monday to give a statement to parliament on the European Union. [Sentence 2] Blair's office said his Foreign Secretary Jack Straw would take his place on Monday to give a statement to parliament on the EU meeting the prime minister attended last week.\\\vspace{0.1cm}
& (QNLI, Not Entailment) & [Question] What is the main executive body of the EU? [Sentence] This means that the Commission has a monopoly on initiating the legislative procedure, although the Council is the "de facto catalyst of many legislative initiatives". \\
\bottomrule
\end{tabular}
}
\caption{Examples of regression errors present when upgrading to the vanilla new model $f^{o}_{new}$ but fixed by our Gated Fusion approach, i.e. predictions of $(f_{old}, f^{o}_{new}, f^{*}_{GF})$ are \textit{(correct, incorrect, correct)}, respectively.}
\label{tab:examples}
\end{table*}

\subsection{Limitations of New Model Ensemble}

In previous works \citep{yan2021positive,work-in-progress}, new model ensemble via majority voting is shown to effectively reduce negative flips and posed as a difficult-to-beat baseline.
Here, we increase the number of models in ensemble to examine its limitations.
Results in Table \ref{tab:ensemble} show that ensemble with more models generally help to obtain lower \nfr.
However, \nfr converges quickly as number of models increased, where a notable gap remains between new model ensemble and Gated Fusion.
Moreover, the results show once more that boosting accuracy does not necessarily improve the backward compatibility in model upgrade. 

In principle, two sources could cause negative flips during model upgrade (a) the stochasticity during model training, including initializations, data loading order, and optimization process \citep{somepalli2022can}.
(b) the distinctions between old and new model hypotheses, including architecture and pretraining data and procedure, leading to different representation space structures and prediction behaviors in terms of decision boundaries.
Without an explicit connection to $f_{old}$, new model ensemble can only reduce negative flips primarily caused by the first factor, while our proposed Gated Fusion directly learns to mitigate regression errors regardless of their causes.

Besides, as large-scale generative models become more and more powerful and popular \citep{raffel2020exploring,brown2020language,su2021multi}, it would be difficult to fine-tune them multiple times on a target task for ensemble. 

\subsection{Analysis of Gated Fusion}

Comparing $f^{o}_{new}$ with $f^{*}_{GF}$, we can calculate the \textit{fix rate} and \textit{new fault rate} of our Gated Fusion method.
During an upgrade, if there are $20$ negative flips with $f^{o}_{new}$ and $16$ out of them can be mitigated by $f^{*}_{GF}$, we obtain the fix rate to be $16/20=80\%$.
Similarly, if $f^{*}_{GF}$ introduces another $4$ new negative flips which are not present with $f^{o}_{new}$, the new fault rate is computed to be $4/20=20\%$.
We calculate the $5$-seed average of these two rates across different classification tasks and upgrade scenarios.
In \bert{base} to \bert{large}, the averaged fix rates by Gated Fusion are $68.4\%$ on SST-2, $83.8\%$ on MRPC, and $82.9\%$ on QNLI, with new fault rates being $4.1\%$ on SST-2, $11.3\%$ on MRPC, and $9.7\%$ on QNLI.
In \bert{base} to \electra{base}, Gated Fusion achieves the averaged fix rates $58.0\%$ on SST-2, $50.8\%$ on MRPC, and $75.6\%$ on QNLI, with new fault rates being $2.8\%$ on SST-2, $15.2\%$ on MRPC, and $4.0\%$ on QNLI.
These results show that, on average, Gated Fusion is able to eliminate $69.9\%$ of total regression errors while adding only $7.9\%$ new ones, comparing with doing model upgrade without any treatment, i.e. $f^{o}_{new}$.

Table \ref{tab:examples} shows a few regression error cases fixed by our proposed approach.
In general, Gated Fusion can mitigate negative flips happened on different classes across diverse tasks as well as on inputs with variable lengths.
With closer inspections of $f^{*}_{GF}$, we found that when $f_{new}$ produces incorrect predictions and $f_{old}$ gives correct outputs, $g_\theta$ is capable of putting larger weights on $f_{old}$ to ensure the backward compatibility.
We also observed that the gate $g_{\theta}$ is more prone to over-fitting when the downstream tasks have smaller training set, e.g. MRPC, or are more difficult in nature, e.g. single-sentence task SST-2 versus sentence-pair tasks, which causes Gated Fusion to introduce more new errors, i.e. higher new fault rates.

\section{Discussion}
Gated Fusion requires to host both old and new models at inference time, which could raise a concern regarding the increased computational burden.
However, in practice, old model’s logits of previous inference instances can be cached in storage and later leveraged in our Gated Fusion.
That is, we only need to host the new model with the gate at inference time and leverage old predictions from cache.
And for the out-of-cache inputs, backward-compatibility would be less of an issue since users have not observed such examples to make conclusions on the underlying regression.

For real-world applications, there could be multiple model updates and thus multiple legacy versions.
We note that in this scenario, user experience would be primarily grounded on predictions of the latest legacy version, which are also saved in cache.
Our Gated Fusion can hence leverage them and make new model’s predictions compatible to those from the latest legacy version.

In addition, we emphasize that the main challenge in the regression reduction research problem is to find the best trade-off between model effectiveness and backward compatibility.
In this work, we show that the weighted ensemble of old-new models with a learned gate, which we call Gated Fusion, achieves a better negative flip rate than previously explored methods for regression reduction, while straight-forward ensemble approaches cannot naturally weigh on this trade-off.
We don’t claim to invent the gated ensemble of old and new models but rather that our main contribution is to show that by repurposing the classic gating mechanism, the gated ensemble can become the most competitive approach to the challenging model-upgrade regression reduction problem, with no overall performance degradation on two realistic model update scenarios across three different datasets.

Recently, more and more NLP products have been deployed in the industry as this field matures.
We would like to stress that as better NLP models are being developed, the backward-compatible model upgrade problem naturally emerges as the new research topic strongly motivated by the real-world challenges. 
While backward-compatibility is currently a niche research topic, we believe that there are many thrilling future directions worth to be investigated.

\section{Related Work}

\citet{yan2021positive} first studied the backward compatibility of predictions during model upgrade on image classification tasks.
Later, \citet{work-in-progress} investigated the similar topic in natural language understanding and formulated it as a constrained optimization problem.
They both show that customized variants of knowledge distillation \citep{Hinton2015DistillingTK}, which align the predictions of old and new models on potential regression errors, are effective approaches.
A model ensemble has also shown to be surprisingly effective \citep{yan2021positive,work-in-progress}, despite no explicit connection between old and new models.
This was credited to variance reduction in model predictions, making it less prone to over-fitting and reducing regression errors indirectly.
In this work, we leverage the gating mechanism to combine old and new models to further reduce model upgrade regression errors by a large margin across classification tasks.

\citet{cai2022measuring} analyzed and proposed backward congruent re-ranking to reduce regression in model upgrades for structured predictions tasks such as dependency parsing and conversational semantic parsing.
\citet{trauble2021backward} proposed an efficient probabilistic approach to locate data instances whose old predictions could be incorrect and update them with ones from the new model.
\citet{zhou2022forward} looked into forward compatibility, where new classes can be easily incorporated without negatively impacting existing prediction behavior.
More recently, \citet{schumann2023backward} inspected classification model regression during training data updates and mitigated the problem by interpolating between weights of the old and new models.
On top of that, learning cross-model compatible embeddings has been extensively explored in visual search \citep{chen2019r3,hu2019towards,wang2020unified}.
Several techniques have been proposed to optimize cross-model interoperability of embeddings, including metric space alignment \citep{Shen2020TowardsBR}, architecture search \citep{duggalcompatibility}, and aligning class centers between models \citet{meng2021learning}.
In this work, we focus on improving backward compatibility during model upgrade in terms of prediction behavior on classification tasks, i.e. old and new models should produce consistently correct predictions.

Reducing regression during model upgrade is also related to continual learning \citep{parisi2019continual,de2019continual,sun2019lamol,chuang2020lifelong,sachidananda2021efficient}, incremental learning \citep{chaudhry2018riemannian,shan2020learn} and concept drifting \citep{gama2014survey,vzliobaite2016overview,ganin2016domain,zhuang2020comprehensive,lazaridou2021pitfalls}.
In these problems, models are required to learn from and deal with continuously changing data (in terms of examples, classes or tasks), and also need to prevent the forgetting of previously learnt knowledge.
This could be one potential cause of regression observed at inference. 
However, in backward-compatible model upgrade, a new model, usually with distinct network architecture, is trained from scratch to perform the same task and is expected to behave similarly wherever the previous model predicts correctly.

The gating mechanism is widely adopted by recurrent neural networks to effectively control information flows across networks \citep{hochreiter1997long,cho2014properties,van2016pixel,dauphin2017language,lai2019goal} and contextualize embeddings \cite{peters2018deep,lai2020context}.
It is then repurposed to act as a switch for the mixture of different prediction modes, notably to combine input word copying based on the pointer network \citep{vinyals2015pointer} with the word generation from output vocabulary \citep{gu2016incorporating,merity2016pointer,see2017get}.
Our proposed approach is inspired by these works and leverages the gating mechanism to effectively combine old and new models to improve backward compatibility during model upgrade.


\section{Conclusion}
Ensuring backward compatibility during model upgrade has become a critical topic in real-world NLP applications.
In this work, we proposed a new approach, \emph{Gated Fusion}, that achieves significantly better backward compatibility without compromising accuracy performance on two challenging upgrade scenarios for NLP classification.
Experiments demonstrated that our approach outperforms competing methods and achieves negative flip rate reductions by up to $73.2\%$.
Our future research includes improving backward compatibility beyond classification to span detection, model upgrades with very large language models, and upgrades on training data or label schema.
We hope that this work can inspire further research and make progress towards smoother transitions of prediction powers as NLP systems evolve.

\section*{Limitations}
Our proposed method mostly works on the upgrades of underlying pretrained language models for NLP classification tasks.
Potential limitations include applying our approach on distant tasks such as question answering or information retrieval, upgrade to models from different architecture families such as recurrent neural nets, and the inapplicability of our method to more recent learning formulation such as in-context learning via prompting. 

\section*{Ethics Statement}
Prediction backward compatibility during model upgrade is an emerging research topic to ensure positive congruency and smoother transitions from existing models towards more performant systems.
With primary evaluation on accuracy and negative flips, we acknowledge that our method may also inherit social biases and other toxicity persisted in the legacy models.
On the other hand, we have noted that fairness and safety have been one of principal criteria when developing system upgrades.
Investigations of the inheritance of persistent toxicity and mitigation of it during backward-compatible upgrades merit interests of future research.

\section*{Acknowledgements}
We would like to acknowledge AWS AI Labs for inspiring discussions, honest feedback, and full support. 
We are also very grateful to reviewers for judicious comments and valuable suggestions.

\bibliography{anthology,custom}
\bibliographystyle{acl_natbib}

\clearpage
\appendix

\section{Details on Experiment Settings}

\subsection{Model Training Hyper-parameters}

We search among following hyper-parameter space for the training of the old model $f_{old}$ and the new model in the vanilla setting $f^{o}_{new}$ across all datasets:
\begin{itemize}[itemsep=0pt,topsep=0pt,parsep=0pt,partopsep=0pt]
    \item Learning Rate: $5e^{-6}$, $1e^{-5}$, $3e^{-5}$, $5e^{-5}$
    \item Batch Size: $16$, $32$
    \item Training Epochs: $3$, $5$, $8$.
\end{itemize}
The selected hyper-parameters for each model with \textit{(learning rate, batch size, training epoch)}:
\begin{itemize}[itemsep=0pt,topsep=0pt,parsep=0pt,partopsep=0pt]
    \item \bert{base}:
    \begin{itemize}
        \item On SST-2: $(\text{lr }1e^{-5}, \text{batch }16, \text{epoch } 5)$
        \item On MRPC: $(\text{lr }3e^{-5}, \text{batch }16, \text{epoch } 5)$
        \item On QNLI: $(\text{lr }3e^{-5}, \text{batch }32, \text{epoch } 3)$
    \end{itemize}
    \item \bert{large}:
    \begin{itemize}
        \item On SST-2: $(\text{lr }1e^{-5}, \text{batch }16, \text{epoch } 5)$
        \item On MRPC: $(\text{lr }3e^{-5}, \text{batch }16, \text{epoch } 5)$
        \item On QNLI: $(\text{lr }3e^{-5}, \text{batch }32, \text{epoch } 3)$
    \end{itemize}
    \item \electra{base}:
    \begin{itemize}
        \item On SST-2: $(\text{lr }1e^{-5}, \text{batch }16, \text{epoch } 5)$
        \item On MRPC: $(\text{lr }5e^{-5}, \text{batch }32, \text{epoch } 5)$
        \item On QNLI: $(\text{lr }3e^{-5}, \text{batch }32, \text{epoch } 3)$
    \end{itemize}
\end{itemize}

These model training hyper-parameters for a specific model on one specific dataset is then fixed and reused for all the competing methods to improve backward compatibility during model upgrade.

\subsection{Distillation Hyper-parameters}

The knowledge distillation method from \citet{work-in-progress} imposed an additional loss $\lambda \cdot KL(l_{old} / T, l_{new} / T)$ on potential regression instances.
We experimented the best possible hyper-parameters from the following:
\begin{itemize}[itemsep=0pt,topsep=0pt,parsep=0pt,partopsep=0pt]
    \item $\lambda$: $0.1, 1.0, 10.0$
    \item Temperature $T$: $0.5, 1.0, 2.0$
\end{itemize}

\subsection{Details on Gated Fusion}
We initialize the gate $g_{\theta}$ to be a two-layer feed-forward network with the architecture \textit{(Dropout, Linear, LayerNorm, ReLU, Dropout, Linear, Sigmoid)} and fix the hidden size to be $64$ across all our experiments.

During the training of Gated Fusion, we only train the $f_{new}$ within $f^{*}_{GF}$ for the first $(N - 1)$ epochs to ensure its competence, where $N$ is the total training epochs.
In the last training epoch, we jointly train $g_{\theta}$ and $f_{new}$ using the Gated Fusion logits $l^{*}_{GF}$ with the secondary learning rate $lr2$.
To prevent the overfitting of the gate, we also apply \textit{drop\_gate} where at each training step during the last epoch, there is $D\%$ to only train $f_{new}$ within $f^{*}_{GF}$ and $(1 - D)\%$ to train with $l^{*}_{GF}$.

The hyper-parameter space of Gated Fusion is listed as follows:
\begin{itemize}[itemsep=0pt,topsep=0pt,parsep=0pt,partopsep=0pt]
    \item Drop Gate ($\%$): $40, 50, 60, 80$
    \item Temperature $T$ on old logits: $1.0, 1.2, 1.4, 1.6$
    \item lr2: $5e^{-7}$, $1e^{-6}$, $3e^{-6}$, $1e^{-5}$, $3e^{-5}$
\end{itemize}
We found that to achieve good results, the gap in logit magnitude of $f_{old}$ and $f_{new}$ needs to be bridged by the temperature when upgrading from \bert{base} to \electra{base}, with $T$ being $1.6$ on SST-2, $1.6$ on MRPC, and $1.2$ on QNLI.
On the other hand, $T=1$ gives good results across three datasets when upgrading from \bert{base} to \bert{large}.
This could result from the distinct pretraining schemes between models where MLM seem to produce larger magnitude of output logits.

\end{document}